\documentclass{article}
\usepackage{ijcai24}

\usepackage{times}
\usepackage{soul}
\usepackage{url}
\usepackage[hidelinks]{hyperref}
\usepackage[utf8]{inputenc}
\usepackage[small]{caption}
\usepackage{graphicx}
\usepackage{amsmath}
\usepackage{amsthm}
\usepackage{booktabs}
\usepackage{algorithm}
\usepackage{algorithmic}
\usepackage[switch]{lineno}
\usepackage{xcolor}
\usepackage{multicol}
\usepackage{multirow}
\usepackage{subcaption}
\usepackage{enumitem}
\usepackage{cleveref}
\usepackage{epsfig}
\usepackage{graphicx}
\usepackage{amsmath}
\usepackage{amssymb}
\usepackage{newfloat}
\usepackage{booktabs}
\usepackage{array}
\usepackage{xspace}
\usepackage{bbm}
\usepackage{tabularx}

\newcolumntype{Y}{>{\centering\arraybackslash}X}

\newcommand{\model}{SynFormer3D\xspace}
\newcommand{\dataset}{SynVL3D\xspace}

\title{3D Vision and Language Pretraining with Large-Scale Synthetic Data}

\author{
Dejie Yang$^{1}$, 
Zhu Xu$^{1}$ ,
Wentao Mo$^{1}$ , 
Qingchao Chen$^{2,3}$, 
Siyuan Huang$^{4}$ , 
Yang Liu$^{1,3}$\footnote{Corresponding Author}\\\
\affiliations
$^1$ Wangxuan Institute of  Computer Technology, Peking University\\
$^2$ National Institute of Health Data Science, Peking University\\
$^3$ National Key Laboratory of General Artificial Intelligence, Peking University\\
$^4$State Key Laboratory of General Artificial Intelligence, BIGAI\\
\emails
\{ydj, xuzhu\}@stu.pku.edu.cn, huangsiyuan@ucla.edu, \{mowt, qingchao.chen, yangliu\}@pku.edu.cn}

\begin{document}

\maketitle

\begin{abstract}
3D Vision-Language Pre-training (3D-VLP)  aims to provide a pre-train model which can bridge 3D scenes with natural language, which is an important technique for embodied intelligence.  However, current 3D-VLP datasets are hindered by limited scene-level diversity and insufficient fine-grained annotations (only 1.2K scenes and 280K textual annotations in ScanScribe), primarily due to the labor-intensive of collecting and annotating 3D scenes. To overcome these obstacles, we construct SynVL3D, a comprehensive synthetic scene-text corpus with 10K indoor scenes and 1M descriptions at object, view, and room levels, which has the advantages of diverse scene data, rich textual descriptions, multi-grained 3D-text associations, and low collection cost. Utilizing the rich annotations in SynVL3D, we pre-train a simple and unified Transformer for aligning 3D and language with multi-grained pretraining tasks. Moreover, we propose a synthetic-to-real domain adaptation in downstream task fine-tuning process to address the domain shift. Through extensive experiments, we verify the effectiveness of our model design by achieving state-of-the-art performance on downstream tasks including visual grounding, dense captioning, and question answering. Codes are available at: \url{https://github.com/idejie/3DSyn}.

\end{abstract}

\section{Introduction}
\label{sec:intro}

\begin{figure}[!t]
    \centering
    \includegraphics[width=1\linewidth]{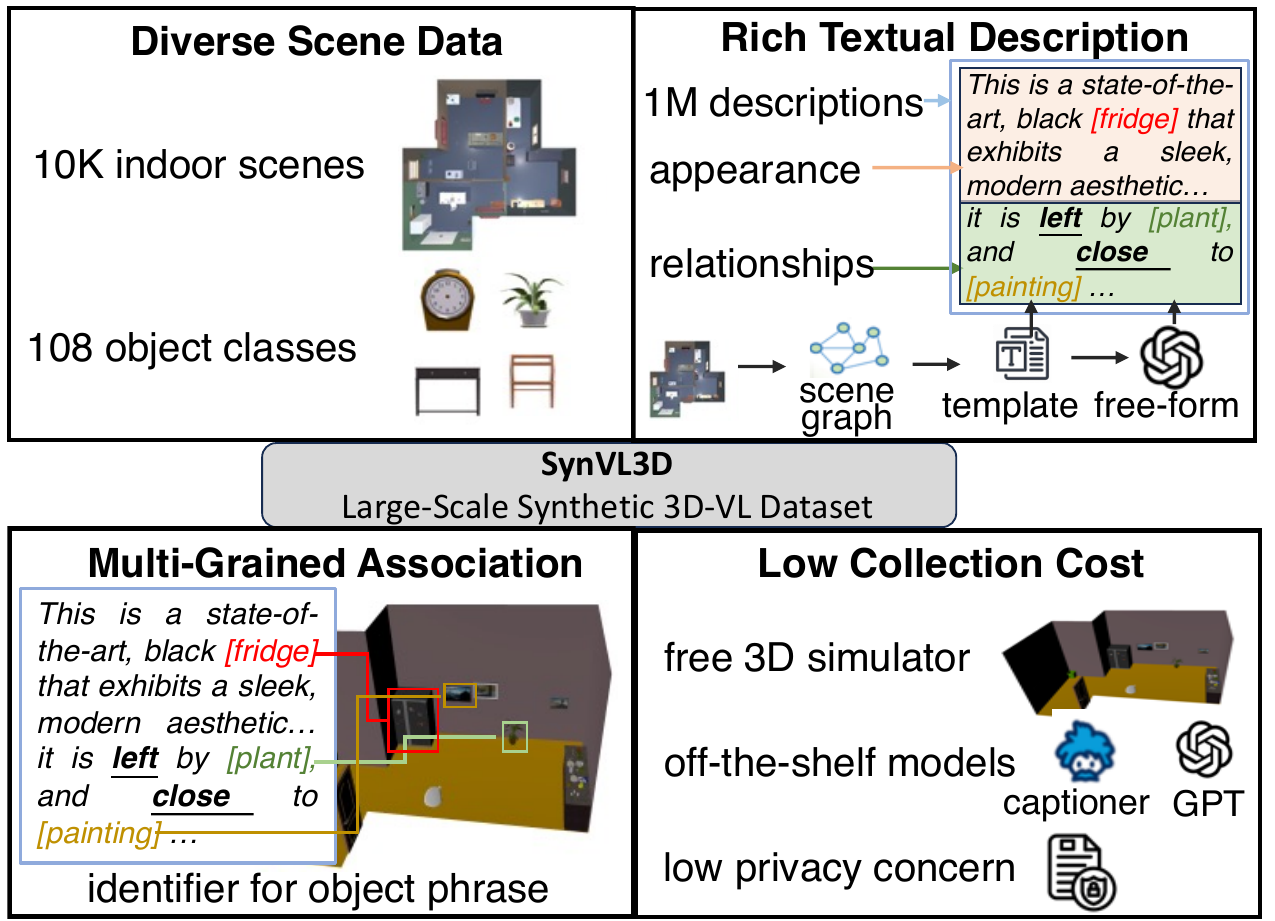}
    \caption{Advantages of our proposed dataset \dataset.
    } 
    \label{fig:overall}
\end{figure}

Bridging 3D scenes with natural language represents a pivotal advancement in the pursuit of embodied artificial intelligence~\cite{huang2023embodied} including  3D vision-language (3D-VL) tasks such as 3D visual grounding~\cite{scanrefer,referit3d}, dense captioning~\cite{scan2cap}, {and} question answering~\cite{scanqa,3DVQA_2024_AAAI},etc. 

Drawing inspiration from the impact of 2D vision-text pre-training models~\cite{radford2021learning,flamingo}, recent researches~\cite{zhu20233d,jin2023context} demonstrate that  3D-VL  pre-training can enhance the performance in subsequent tasks. 
However, the performance of these methods is limited by the size and diversity of 3D-text data, as 3D data collection largely dependents on the scanning device, introducing inherent complexities and increased costs in contrast to the simpler process of gathering 2D images.
There are limitations to the existing 3D-VL datasets:
(1) \textit{limited diversity of 3D vision}, the largest-scale 3D-VL dataset ScanScribe~\cite{zhu20233d} combines multiple existing 3D datasets, but still only contains 1.2K indoor scenes, which causes limited vision diversity for pre-training; 
(2) \textit{insufficient fine-grained 3D-text} associations, ScanScribe provides 278K descriptions of objects where only the sentence-level target objects is annotated and ignores the other fine-grained vision-language associations between the spatial region and other objects mentioned in the sentence.
(3) \textit{labor-intensive data collection procedure},
the 3D data collection of a real scene needs 30 minutes with advanced scanners~\cite{Yeshwanth_2023_ICCV} and annotating a scene with the help of the automatic segmentation still requires 1.93 minutes, makes large-scale real-world 3D scene infeasible to collect.

To address the limitation of existing 3D-VL datasets, we propose the dataset \dataset, which contains the free-collected and easy-to-annotate synthetic 3D data generated by a 3D scene simulator~\cite{deitke2022️}, instead of using realistic data. 
As shown in Figure \ref{fig:overall}, our dataset \dataset has four advantages: 
(1) \textbf{Diverse scene data}. Benefit {from} the capability of the 3D simulator~\cite{deitke2022️}, we collected 10,000 indoor 3D scenes with 108 object categories.
And \dataset provides semantic (category), visual (high-quality mesh), and fine-grained visual annotation (error-free position, orientation and segmentation mask) of each object .
To acquire fine-grained understanding for scenes, we formed a scene graph for each house (scene) with the relationships between objects based on their spatial position, size, and shape.
(2) \textbf{Rich textual description}. We utilize templates, off-the-shelf image captioners~\cite{dai2023instructblip}, and GPT3~\cite{gpt-3}
to generate appearance and relationship descriptions of objects,  and obtain more than 1M {template-based or free-form} descriptions. 
(3) \textbf{Multi-grained association}.
{Based on the above textual descriptions and scene graph,  we retrain a distinct identifier for every noun phrase, creating a connection between each identifier and a spatial region. This establishes a robust phrase-region association, enhancing the grounded interpretation and facilitating a more meaningful evaluation.}
(4) \textbf{Low collection cost}. Based on simulators, we can obtain free 3D scene data, use off-the-shelf models to automatically describe scenes and obtain fine-grained annotations with few manual annotations. 
With virtual scenes, we also avoid privacy issues compared to real scenes.

Based on the diverse and fine-grained annotations of large scale of 3D scene provided by \dataset, we can construct more \textbf{ fine-grained pre-training tasks}. The tasks include relationship prediction between objects, multi-level and view-aggregated region-word alignment, beyond scene-text matching and masked language/object {modeling} that previous pre-training methods~\cite{zhu20233d,jin2023context} use.
The multi-grained 3D VLP tasks enable the model to capture rich 3D vision-language knowledge, which benefits downstream {vision-language} tasks. 
However, pre-training based on synthetic 3D data suffers from the domain shift between the synthetic data and the realistic data used in downstream tasks.
Thus, we introduce \textbf{synthetic-to-real adaptation}  with domain discriminator in the fine-tuning process of downstream tasks to reduce the representation distribution difference between the synthetic and realistic data of our pre-trained model.
The efficacy of our model is evaluated across 3D-VL tasks, including visual grounding (e.g., ScanRefer~\cite{scanrefer}, Nr3D/Sr3D~\cite{referit3d}), dense captioning (e.g., Scan2Cap~\cite{scan2cap}), question answering (e.g., ScanQA~\cite{scanqa}).

{Our main contributions can be summarized as follows:
(1) We construct a large-scale synthetic dataset \dataset, featuring diverse 3D scenes, textual descriptions, and fine-grained 3D region-text associations at a low manual cost, which is a significant improvement in terms of data diversity and scale compared to prior datasets.
(2) We introduce a pretraining framework featuring new auxiliary tasks such as pair-wise object relationship prediction and multi-level 3D region-phrase alignment to enhance comprehensive understanding. To address the domain shift between synthetic and real 3D data, we introduce a domain adaptation technique, improving the fine-tuning process for downstream tasks.
(3) We demonstrate that with the data scale-up and model design, our model \model achieves state-of-the-art performance on a variety of 3D-VL tasks, including visual grounding, dense captioning, and question answering.}

\section{Related Work}
\label{sec:related}

\subsection{3D Vision-Language Dataset}
There has been a growing interest in 3D-VL datasets. ~\cite{scanrefer} and ~\cite{referit3d} introduced the ScanRefer and ReferIt3D datasets, which serve as benchmarks for anchoring sentence-level natural language to ground the 3D object. ~\cite{abdelreheem2024scanents3d} and ~\cite{yuan2022toward} extend to provide phrase-level grounding annotations.
Scan2Cap~\cite{scan2cap} provides the textual descriptions about the scene in ScanRefer for captioning.
ScanQA \cite{scanqa} developed a 3D question-answering dataset, which annotates the objects mentioned within the questions and answers. 
ScanScribe~\cite{zhu20233d} endeavors to build a large-scale dataset by combining the aforementioned datasets for pre-training purposes. 
However, these 3D-VL datasets are constrained by a lack of scene-level diversity, fine-grained annotations (the largest existing dataset ScanScribe with only 1.2K scenes and 280K textual descriptions, and the laborious process involved in collecting and annotating 3D scenes.  To overcome these obstacles, we constructed the dataset \dataset, which features freely collected and easily annotated 3D data generated by a 3D scene simulator ~\cite{deitke2022️}, as opposed to utilizing realistic data. It encompasses diverse 3D scene data (10K scenes), provides 1M rich textual descriptions with low cost. 
Moreover, compared to the sentence-region association in ScanScribe, we provide the phrase-region finegrained association, which means that most noun phrase in description has an association with the spatial region in the 3D scene.

\subsection{Large-scale 3D-VL Pre-training}
In recent years, large-scale pre-training has emerged as a fundamental paradigm in natural language processing (NLP)~\cite{gpt-1,gpt-2,gpt-3} and 2D vision-and-language (2D-VL) domains~\cite{lu2019vilbert,radford2021learning}. While the pre-training technique has become indispensable in these domains, its application in the 3D-VL field remains relatively uncharted. 
3D-VISTA~\cite{zhu20233d} seeks to pioneer a pre-training model for 3D-VL tasks, employing strategies such as masked language modeling, masked object modeling, and scene-sentence matching. Concurrently, 3D-VLP~\cite{jin2023context} introduces a context-aware spatial-semantic alignment to forge connections between point clouds and the textual descriptions of ground objects. However, the efficacy of these pre-training tasks is often compromised by the lack of detailed annotations in existing datasets.
In this paper, we propose three fine-grained pre-training tasks, including object relations prediction, multi-level region word alignment and view-aggregated region word alignment, to enhance the model's perception capability of object relationships, scenes at different levels and from various views.

\subsection{Simulation-to-Real Domain Adaptation}

Simulation-to-real domain adaptation refers to the process of transferring knowledge acquired in a simulated environment to the real world, and is widely applied in 2D vision~\cite{pei2024evidential,zhao2023da,pei2023uncertainty,xu2022cda,chen2020structure,chen2018re} and 3D fields like autonomous vehicles~\cite{zhao2021epointda}, indoor scenes layout~\cite{ding2022doda}, and robotics~\cite{debortoli2021adversarial}.
Existing methods primarily aim to design a simulation-to-real domain adaptation framework for uni-modal applications, such as the 3D semantic segmentation ~\cite{zhao2021epointda,ding2022doda},  point cloud object detection~\cite{debortoli2021adversarial} and point cloud registration~\cite{chen2023sira}.
However, there are few works exploring the simulation-to-real domain adaptation in 3D cross-modality applications, where the domain shift exists in both 3D vision and language.
To address this dual and joint domain shift, we propose a task-specific fine-tuning strategy with domain adaptation in 3D vision, language, and joint 3D-language domains.
\section{Dataset Generation: \dataset}
\label{sec:dataset}

\begin{table}
\centering
\resizebox{\linewidth}{!}{\begin{tabular}{ccccc}
\toprule
\textbf{Dataset} &  \textbf{Type}& \textbf{Text} & \textbf{Scene} & \textbf{Annotation}\\
\midrule
Nr3D~\cite{referit3d} &Realistic & 31K &707 &Sentence-Region \\
ScanRefer~\cite{scanrefer} &Realistic & 37K &800&Sentence-Region \\
ScanQA~\cite{scanqa}  &Realistic& 41K &800&Sentence-Region  \\
ScanEnts3D~\cite{abdelreheem2024scanents3d}&Realistic&84K&705&Phrase-Region\\
PhraseRefer~\cite{yuan2022toward}&Realistic&88K&707&Phrase-Region\\
3R-Scan~\cite{3rscan}&Realistic&90K&478&Sentence-Region\\
Sr3D~\cite{referit3d}&Realistic & 91K &707 &Sentence-Region \\
ScanScribe~\cite{zhu20233d} &Realistic & 278K & 1.2K&Sentence-Region\\
\midrule
Syn3D(Ours) &Synthetic & 1M & 10K&Phrase-Region\\
\bottomrule
\end{tabular}}
\caption{The comparison between \dataset and other 3D-VL datasets.  ``Type'' stands for the source of 3D scenes. ``Text'' denotes the textual description size. 
} 
\label{dataset}
\end{table}
In this paper, we introduce \dataset, the first large-scale dataset of 3D synthetic scene-text pairs designed for 3D-VL pre-training. IAs indicated in Table \ref{dataset}, \dataset not only surpasses existing 3D-VL datasets in size but also offers a richer variety of 3D scenes, text, and fine-grained 3D-text annotations.
Our dataset \dataset showcases key advantages in four aspects:

\paragraph{Diverse 3D Scene Data.}
We collected mesh data for 10,000 distinct scenes from an off-the-shelf 3D scene simulator~\cite{deitke2022️}. It encompasses 108 object categories, totaling over 1 million distinct instances. Besides the variety surpasses the existing datasets as shown in Table \ref{dataset}, we provide following annotation: 
(1) For each \textit{Single Instance}, we automatically acquire its semantic category and precise bounding boxes and segmentation masks directly from the object and scene information provided by the simulator~\cite{deitke2022️} at no extra cost.
(2) For \textit{Multiple Instance} relationship, we construct a scene graph for each scene based on densely annotated objects and their attributes, where each node represents an object instance, and edges represent the relationships between an instance and its neighbors. 
Inspired by \cite{wald2020learning}, the scene graph encompasses a rich array of relationships, which can be classified into three types: 
a) support relationships(e.g., standing, lying);
b) spatial relationships (e.g., next to, in front of) that consider the relative positions of neighboring instances; and c) comparative relationships (e.g., bigger than, same shape as) that are based on the size and shape of an instance.

\paragraph{Rich Textual Description.}
We have generated more than 1 million textual descriptions, both templated and free-form, encapsulating detailed appearances and relationships between objects.
(1) \textit{Template-based:}
For object appearances, we utilize an off-the-shelf image captioner~\cite{dai2023instructblip} to generate captions for the front view of each object. We manually reviewed these descriptions and make corrections where necessary.
For relationship triplets of the form $<$\textit{object1, relation, object2}$>$, we craft descriptions using the template ``the \textit{object1} is \textit{relation} to \textit{object2}". The description of an instance is structured as a paragraph by merging its appearance description with the relationship descriptions of its  neighbors. 
(2) \textit{Free-form:}
We leverage GPT-3~\cite{gpt-3} to rephrase the templated description to enhance the naturalness. These descriptions not only enrich the dataset with diverse visual perspectives but also enhance the semantic understanding of spatial relations and object attributes.

\paragraph{Phrase-Region Association.}
{The grounding annotations in previous datasets~\cite{referit3d,scanrefer,zhu20233d} mainly consist  sentence-region association for the single referred target object.
In this paper, for 10K scene, we aim to curate grounding annotations of most objects in each description by providing phrase-region explicit association with an identifier of a phrase that refers to an object in a 3D scene. In practice,}
GPT-3 is tasked with rephrasing the descriptions while maintaining a critical requirement: the preservation of the identifier for each object, formatted as \textit{category(id of instance)}, throughout the rephrasing process. This ensures that the region-word association remains intact, maintaining the accuracy of object referencing in the dataset.

\paragraph{Low Collection Cost.} Unlike other datasets in Table \ref{dataset} that are manually collected and annotated from real-world scenes, our \dataset leverages freely synthetic 3D scene data. We utilize off-the-shelf models for automatic scene description and require minimal manual annotation cost. By using virtual scenes, our approach also circumvents privacy issues that can arise from scanning real-world environments.

\section{Methodology}
\label{sec:method}
\begin{figure*}[!ht]
    \centering
    \includegraphics[width=0.9\linewidth]{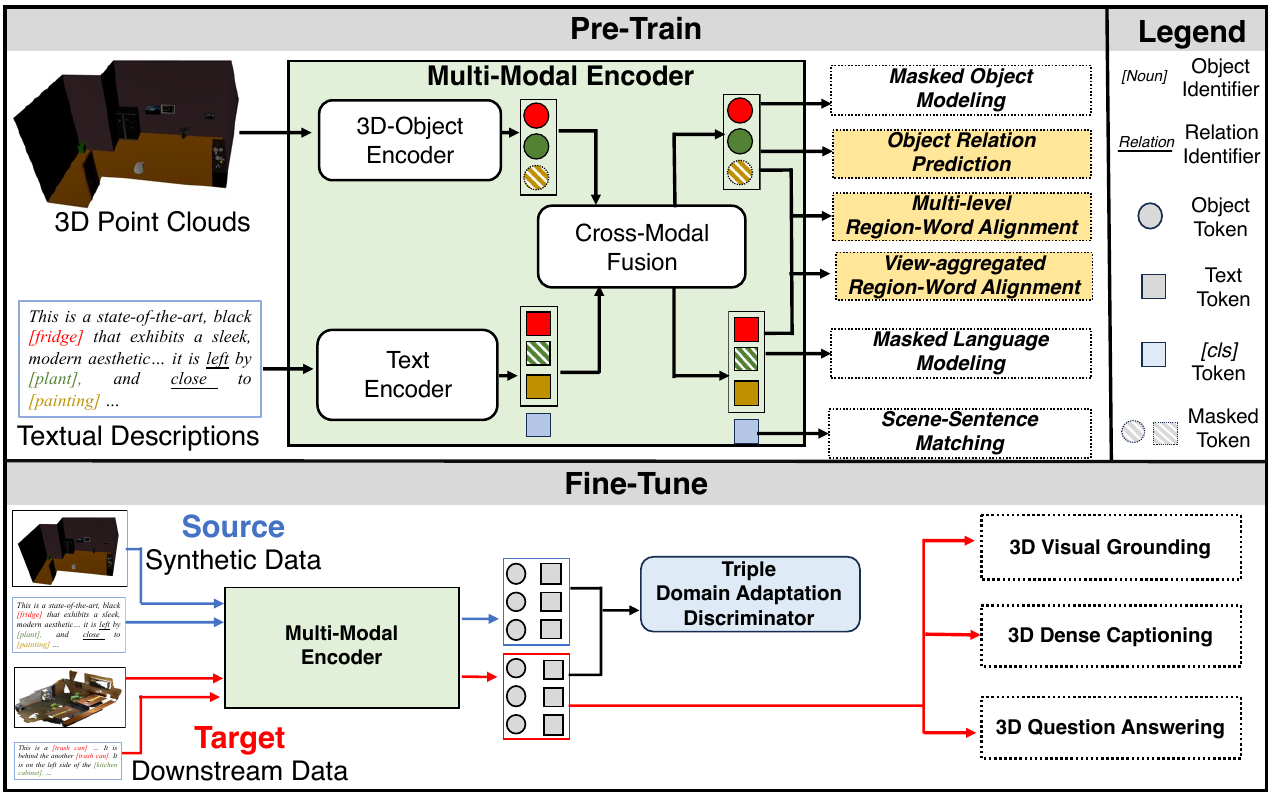}
    \caption{The model architecture of our \model. The multi-modal encoder includes a 3D-object encoder, text encoder, and cross-modal fusion modules. Compared to previous pre-trained models, our \model introduces more {fine-grained} auxiliary pre-training tasks, which include Object Relation Prediction {,} Multi-level and View-aggregated Region-Word Alignment. 
    }
    \label{fig:model}
\end{figure*}
As shown in Figure \ref{fig:model}, our \model first uses multi-modal encoder to extract and fuse 3D and text features. The pipeline mainly consists two steps: synthetic pre-training (Section \ref{pretrain}) and task-specific fine-tuning (Section \ref{finetune}).

\subsection{Multi-Modal Encoder}\label{encoder}
Following \cite{zhu20233d}, firstly we employ text encoder to encode words to text tokens $\{ \mathbf{w}_{\text{CLS}},\mathbf{w}_{1:L} \}$, where $\mathbf{w}_{\text{CLS}}$ denotes the $[cls]$ token and  $L$ denotes the length of a description.
The 3D-object encoder is  to encode  $M$ object proposals {and extract features} from the input point cloud, represented by $\mathbf{o}_{1:M}$.
To achieve multi-modal fusion, we concatenate the text tokens and the 3D object tokens $\{\mathbf{w}_{\text{CLS}}, \mathbf{w}_{1:L}, \mathbf{o}_{1:M}\}$ and forward them to a Transformer.
The outputs of the cross-modal fusion module are denoted as $\{\mathbf{W}_{\text{CLS}}, \mathbf{W}_{1:L}, \mathbf{O}_{1:M}\}$, corresponding to $[cls]$ token, word tokens and object features.

\subsection{Synthetic Pre-Training}
\label{pretrain}
{To align 3D scene and text in self-supervised manner, pre-training tasks Masked Language Modeling (MLM), Masked Object Modeling (MOM), and Scene-Sentence Matching (SSM) are adopted. Further, to achieve better alignment, we propose three fine-grained pre-training tasks  based on the fine-grained annotation in \dataset: Object Relationship Prediction (ORP), Multi-level Region-Word Alignment (MRWA) and View-aggregated Region-Word Alignment (VRWA), details are as follows:}

\paragraph{ORP.} {The previous methods \cite{zhu20233d,jin2023context} neglect the importance of relationship understanding between object pairs, resulting in limitations for downstream tasks. And we introduce object relationship prediction task, which predicting pairwise relationship with prediction head $r(\cdot,\cdot)$, and the relationship annotations in \dataset are utilized to form label.}
{
\begin{equation}
    \mathcal{L}_{\text{ORP}} = -\frac{1}{M^2} \sum_{i=1}^{M} \sum_{j=1}^{M} R_{i,j} \cdot \log(r(\mathbf{O}_i,\mathbf{O}_j)),
\end{equation}
where $i$ and $j$ denote the indices of object features within the set of object proposal features $\mathbf{O}$, $M$ is the total number of object proposal and $R_{i,j}$ is the label of relations between object $i$ and $j$.
}

\paragraph{MRWA.}
{3D tasks and operations involve multi-level targets with varying granularity, ranging from object, room, to entire scene. To achieve multi-level alignment, we align regions and words at object, room and scene levels respectively: }

\begin{equation}
\mathcal{L}(\mathbf{x},\mathbf{z}) = -\frac{1}{L\cdot M}  \sum_{i=1}^{L} \sum_{j=1}^{M}   y_{i,j} \cdot \log (p(\mathbf{x}_i, \mathbf{z}_j)),
\label{match}
\end{equation}
\begin{equation}
\mathcal{L}_{\text{MRWA}} = \sum_{l}\mathcal{L}(\mathbf{W}^l,\mathbf{O}^l),
\end{equation}
where  $L$ and $M$ are the length of text(word) token sequence and the number of object features respectively. $y_{i,j}$ means the binary label of  whether the  $i$-th word $ \mathbf{x}_i$ and $j$-th region(object) $\mathbf{z}_j$  is matching.  
$l \in \{object, room, scene\}$ indicate the levels of a single object, room, and the entire scene, respectively.  $\mathbf{W}^l$ represents the word feature sequence from descriptions at different levels , and $\mathbf{O}^l$ is the object feature set derived from the point cloud of the corresponding level. $p(\cdot,\cdot)$ is the region-word matching prediction head. 
The set of objects involved in different levels of textual description is different. At the object level, only objects near the description are mentioned, while room and house levels provide a larger range of objects.
Therefore, region-word alignment at different levels considers the relationships between objects from the object set at different levels (the self-attention is conducted on different set of objects/words).

\paragraph{VRWA.}  
{The perspective changes when executing operations or object searching in 3D scenes, which makes view-aggregated object-text alignment critical. Following \cite{guo2023viewrefer}, we rotate the point cloud of a 3D scene into $V$ to different views using the annotated camera parameters from \dataset:}
$
\mathbf{P}^v=R(\theta^v) \mathbf {P},
$
where $R(\theta)$ is the rotation matrix~\cite{guo2023viewrefer}, $\mathbf{P}$ is the scene point cloud
and
$\theta^v$ is the rotation angle adopted for $v$-th view: $\theta^v =\frac{2 \pi}{V} (v-1), v \in\{1, \ldots, V\}$
for a total of $V$ views. $R(\theta^v) \times \mathbf{P}$ represents rotating the point cloud of whole scene clockwise by $\theta^v$ degrees along the Z-axis. 
Then we encode the $v$-th view features $\mathbf{O}^v$, and apply multi-modal encoder and aggregate multi-view information for the $i$-th object:
$\mathbf{G}_i = \frac 1 V \sum_{v=1}^{V}\mathbf{O}^v_i.$
 To achieve precise alignment between regions and words (as nouns in descriptions correspond to object identifiers in Figure \ref{fig:model}),
we perform region-phrase matching besides aligning scene and text descriptions:
\begin{equation}
\mathcal{L}_{\text{VRWA}} = -\frac{1}{L\cdot M}  \sum_{i=1}^{L} \sum_{j=1}^{M} z_{i,j}  \log (s( \mathbf{W}_i, \mathbf{G}_j)).
\end{equation}
where $z_{i,j}$ is the matching label of $i$-th word $ \mathbf{W}_i$ and $j$-th region(object) $\mathbf{G}_j$. 
$s(\cdot,\cdot)$ is the region-word matching prediction head.

\paragraph{Final Pre-training Loss.} The final pre-training objective combines the losses from the proxy tasks mentioned earlier:
\begin{equation}
\begin{split}
    \mathcal{L}_{\text{pre-train}} = &\alpha(\mathcal{L}_{\text{MLM}} + \mathcal{L}_{\text{MOM}} + \mathcal{L}_{\text{SSM}})\\
    &+(1-\alpha)(\mathcal{L}_{\text{ORP}} + \mathcal{L}_{\text{MRWA}} + \mathcal{L}_{\text{VRWA}}),
\end{split}
\label{eq:final}
\end{equation}
where $\alpha$ is a hyperparameter used to balance the basic auxiliary tasks(the losses $\mathcal{L}_{\text{MLM}}, \mathcal{L}_{\text{MOM}}, \mathcal{L}_{\text{SSM}}$ is following \cite{zhu20233d}) with our proposed fine-grained pre-training tasks.
Note that the pre-training approach we propose is self-supervised and downstream task-agnostic.
\subsection{Task-Specific Fine-Tuning}
\label{finetune}
By incorporating lightweight task-specific {decoders}, the pre-trained model can be easily adapted to a variety of 3D-VL tasks. 
Due to the possible domain shift in our synthetic dataset and downstream task data (real scene scan data), we propose a task-specific fine-tuning strategy for synthetic-to-real domain adaptation to be better generalized to downstream tasks.
{In particular, we fine-tune the model on the following tasks: 3D Visual Grounding, 3D Dense Captioning, 3D Question Answering. We equip each task a specific lightweight decoder and corresponding fine-tuning losses.} 

\paragraph{Synthetic-to-Real Adaptation.} {Domain shift from multiple perspectives between synthetic data for pre-training and realistic data used in downstream tasks may hinder the performance on various 3D VL tasks. Vision differences can be observed between the synthesized data the scanned real data as shown in Figure \ref{fig:model}. Besides, the descriptions in our dataset are generated by templates or GPT, which also has a gap with artificially annotated natural language descriptions in downstream datasets.
Moreover, downstream tasks also involve the use of joint vision language, such as 3D question answering, where this type of domain shift requires a joint adaptation.}
Therefore, we introduce {synthetic-to-real adaptation}  with a \textit{triple domain discriminator} (shown in Figure\ref{fig:model}) in the fine-tuning process of downstream tasks to reduce the representation distribution difference between the synthetic and realistic data of our pre-trained model.
The triple domain discriminator consists of three separate discriminator for vision, language and vision-language joint domains, addressing unimodal and cross-modal domain shifts.

A vision domain discriminator $D_v$ is placed after multi-modal encoder $E$ to predict the domain label of the object features $\mathbf{O}$: 
\begin{equation}
\mathcal{L}_{\text{vision}}=\max _E \min _{D_v} \mathcal{L}_{BCE}(\mathbf{O})
\label{vision}
\end{equation}
where  $\mathcal{L}_{BCE}$ denotes the binary cross entropy loss.  Gradient Reverse Layers(GRL) is adopted for min-max optimization. 
Similarly, a language domain discriminator $D_l$ is adopted to predict the domain label of sentence tokens $\mathbf{\hat{W}}$: 
\begin{equation}
\mathcal{L}_{\text{lang}}=\max _E \min _{D_l} \mathcal{L}_{BCE}(\mathbf{\hat{W}})
\label{lang}
\end{equation}
where the sentence  token $\mathbf{\hat{W}}$ is the mean of the text token $\mathbf{W}_{1:L}$ in a sentence.
In order to reduce the gap of 3D-text alignment in different domains, we also introduce a joint discriminator $D_{joint}$. And the adaptation loss for shared-domain is: 
\begin{equation}
\mathcal{L}_{\text{joint}}=\max _E \min _{D_{joint}} \mathcal{L}_{BCE}([\mathbf{\hat{W}}_i;\mathbf{O}_i])
\label{joint}
\end{equation}
where $[\mathbf{\hat{W}}_i;\mathbf{O}_i]$ denote $i$-th data sample (setence and corresponding 3D object) which is the concatenation of the $\mathbf{\hat{W}}$ and $\mathbf{O}$.

The final synthetic-to-real adaptation loss can be combined as: $\mathcal{L}_{\text{align}} = \mathcal{L}_{\text{vision}} +\mathcal{L}_{\text{lang}} + \mathcal{L}_{\text{joint}}.$

\paragraph{Final Fine-tune Loss.}The final fine-tuning objective combines the losses from the specific downstream task and  synthetic-to-real domain adaptation:
\begin{equation}
\begin{split}
    {L}_{\text{f}} = &\beta \mathcal{L}_{\text{task}} +(1-\beta) \mathcal{L}_{\text{align}},
\end{split}
\label{eq:fine-tune}
\end{equation}

where $\mathcal{L}_{\text{task}}$ is the specific loss of the downstream task mentioned above and where $\beta$ is a hyperparameter used to balance the downstream task loss and domain-adaptation loss.
\begin{table*}[!ht]
    \centering
   \small
    \begin{tabularx}{\linewidth}{l|ccccc|ccccc|cc}
        \toprule
       \multirow{3}{*}{Method} & \multicolumn{5}{c|}{Nr3D} & \multicolumn{5}{c}{Sr3D} & \multicolumn{2}{c}{ScanRefer}\\
       
         & \multirow{2}{*}{Overall} & \multirow{2}{*}{Easy} & \multirow{2}{*}{Hard} & \multirow{2}{*}{\begin{tabular}[c]{@{}c@{}}View\\ Dep\end{tabular}} & \multirow{2}{*}{\begin{tabular}[c]{@{}c@{}}View\\ Indep\end{tabular}} 
        & \multirow{2}{*}{Overall} & \multirow{2}{*}{Easy} & \multirow{2}{*}{Hard} & \multirow{2}{*}{\begin{tabular}[c]{@{}c@{}}View\\ Dep\end{tabular}} & \multirow{2}{*}{\begin{tabular}[c]{@{}c@{}}View\\ Indep\end{tabular}}&\multirow{2}{*}{\begin{tabular}[c]{@{}c@{}}acc\\ 0.25\end{tabular}}&\multirow{2}{*}{\begin{tabular}[c]{@{}c@{}}acc\\ 0.5\end{tabular}} \\
        & & & & & & & & & & \\
        \midrule

        3DVG-Trans \cite{3dvg} & 40.8 & 48.5 & 34.8 & 34.8 & 43.7 & 51.4 & 54.2 & 44.9 & 44.6 & 51.7 &47.6&34.7\\
        TransRefer3D \cite{transrefer3d} & 48.0 & 56.7 & 39.6 & 42.5 & 50.7 & 57.4 & 60.5 & 50.2 & 49.9 & 57.7 &-&-\\
        LAR \cite{lar} & 48.9 & 58.4 & 42.3 & 47.4 & 52.1 & 59.4 & 63.0 & 51.2 & 50.0 & 59.1 &-&- \\
        SAT \cite{sat} & 56.5 & 64.9 & 48.4 & 54.4 & 57.6 & 57.9 & 61.2 & 50.0 & 49.2 & 58.3 & 44.5 & 30.1  \\
        3D-SPS \cite{3dsps} & 51.5 & 58.1 & 45.1 & 48.0 & 53.2 & 62.6 & 56.2 & 65.4 & 49.2 & 63.2 & 48.8 & 37.0  \\
        MVT \cite{mvt3d} & 59.5 & 67.4 & 52.7 & 59.1 & 60.3 & 64.5 & 66.9 & 58.8 & 58.4 & 64.7& 40.8 & 33.3 \\
        ViL3DRel \cite{vil3dref} & \underline{64.4} & 70.2 & \textbf{57.4} & \underline{62.0} & 64.5 & 72.8 & 74.9 & 67.9 & \underline{63.8} & 73.2 & 47.9 & 37.7 \\ 
        ScanRefer \cite{scanrefer}  & -&-&-&-&-&-&-&-&-&-& 41.2 & 27.4 \\
        3DJCG \cite{3djcg}   & -&-&-&-&-&-&-&-&-&-& 49.6 & 37.3 \\
        \midrule

        3D-VLP\cite{jin2023context}& -&-&-&-&-&-&-&-&-&-&  \underline{51.4} & {39.5} \\
        3D-VISTA\cite{zhu20233d} & 64.2 & \underline{72.1} & \underline{56.7} & 61.5 & \underline{65.1} & \underline{76.4} & \underline{78.8} & \underline{71.3} & 58.9 & \underline{77.3} &50.6&\underline{45.8} \\ 
       Ours & \textbf{65.5} & \textbf{73.2} & {56.3} & \textbf{63.2} & \textbf{66.1} & \textbf{77.9} & \textbf{79.2} & \textbf{73.3} & \textbf{64.1} & \textbf{79.2} &\textbf{52.3}&\textbf{46.2} \\
        \bottomrule
    \end{tabularx}
     \caption{Grounding accuracy (\%) on Nr3D, Sr3D and ScanRefer.  The best and second-best results are in \textbf{bold} and \underline{underlined}.} 
        \label{vg1}
\end{table*}

\begin{table*}
\centering
\small
\begin{tabularx}{\linewidth}{lcYYYY|YYYY}
    \toprule
   & \multirow{2}{*}{Method} & \multicolumn{4}{c|}{@0.25} & \multicolumn{4}{c}{@0.5} \\
    && C & B-4 & M  & R  & C   & B-4  & M & R \\ \midrule
    \multirow{4}{*}{No Pre-train}&Scan2Cap \cite{scan2cap} & 53.7  & 34.3 & 26.1 & 55.0 & 35.2 & 22.4 & 21.4 & 43.5 \\
    &X-Trans2Cap\cite{Yuan_2022_CVPR}&58.8&34.2&25.8&54.1&41.5&23.8&21.9&45.0\\
    &MORE\cite{jiao2022more}&58.9&35.4&26.4&55.4&39.0&23.0&21.7&44.3\\
    &3DJCG \cite{3djcg}   & 60.9  & \underline{39.7} & 27.5 & \textbf{59.0} & 47.7 & 31.5 & 24.3  & 51.8 \\ \midrule
    \multirow{3}{*}{Pre-train}&3D-VLP \cite{jin2023context} &64.1&\textbf{39.8}&27.7&\underline{58.8}&50.0&31.9&24.5&51.5\\
    &3D-VISTA\cite{zhu20233d} &\underline{71.0} & 36.5 & \underline{28.4} & 57.6 & \textbf{66.9} & \underline{34.0} & \underline{27.1} & \underline{54.3} \\ 
    &Ours & \textbf{72.1} & {37.3} & \textbf{28.9} & {58.2} & \underline{64.2} & \textbf{34.6} & \textbf{28.0} & \textbf{54.8} \\ 

    \bottomrule
\end{tabularx}
\caption{Dense Captioning results on Scan2Cap dataset. ``C'' stands for ``CIDEr'', ``B-4'' for ``BLEU-4'', ``M'' for ``METEOR'', and ``R'' for ``ROUGE'', respectively. ``@0.25'' and ``@0.5'' represent the 3D IoU between the predicted and annotated box. The best and second-best results are in \textbf{bold} and \underline{underlined}.}
\label{caption}
\end{table*}

\begin{table*}[!t]
\centering

\small

\begin{tabular}{lccccccc}
\toprule
&Method & \multicolumn{1}{c}{EM@1} & \multicolumn{1}{c}{EM@10} & \multicolumn{1}{c}{BLEU-4} & \multicolumn{1}{c}{ROUGE} & \multicolumn{1}{c}{METEOR} & \multicolumn{1}{c}{CIDEr}\\
\midrule
 \multirow{3}{*}{No Pre-train}&ScanQA \cite{scanqa} & 23.5 / 20.9 & 56.5 / \underline{54.1} & 12.0 / 10.8 & 34.3 / 31.1 & 13.6 / 12.6 & 67.3 / 60.2 \\
& CLIP-Guided \cite{parelli2023clip} &23.9 / 21.4& - / -&14.6 / 11.7&35.2 / 32.4&13.9 / \underline{13.3} &69.5 / \underline{62.8}\\
&Multi-CLIP \cite{delitzas2023multi} &  24.0 / 21.6    &- / -               & 12.7 / \textbf{12.9}          & 35.5 / 32.6      & 14.0 / \textbf{13.4}    & 68.7 / \textbf{63.2}\\
 
\midrule
\multirow{3}{*}{Pre-train}&3D-VLP\cite{jin2023context}& 25.2 / 20.4 & 55.2 / 51.5 & 10.5 / 8.7 & 35.5 / 29.6 & 13.8 / 11.6 & 68.6 / 55.7 \\
&3D-VISTA\cite{zhu20233d} & \underline{27.0} / \underline{23.0} & \underline{57.9} / {53.5} & \underline{16.0} / {11.9} & \underline{38.6} / \underline{32.8} & \textbf{15.2} / {12.9} & \textbf{76.6} / {62.6} \\
&Ours &  \textbf{27.6} / \textbf{24.1} & \textbf{58.3} / \textbf{54.5} & \textbf{16.3} / \underline{12.3} & \textbf{39.2} / \textbf{33.3} & \underline{14.9} / {13.1} & \underline{76.2} / {62.7} \\
\bottomrule
\end{tabular}
\caption{Answer accuracy on ScanQA using object proposals from Mask3D. Each entry denotes ``test w/ object'' / ``test w/o object''. The best and second-best results are in \textbf{bold} and \underline{underlined}. }
\label{tab:scanqa}
\end{table*}

\section{Experiments}
\subsection{Implementation Details}
The pre-training runs for 100 epochs with a batch size of 64 with  NVIDIA A100  GPUs. 
We set the balance hyper-parameters $\alpha, \beta$ as $0.5$ and $0.8$.
We use the AdamW~\cite{loshchilov2019decoupled} optimizer and learning rate is set to $1e^{-4}$. 

\subsection{Downstream Task}
\paragraph{3D Visual Grounding.} 
In Table \ref{vg1}, we evaluate our model on Nr3D, Sr3D~\cite{referit3d}  and ScanRefer~\cite{scanrefer}. 
For Nr3D/Sr3D, we report the results as the grounding accuracy. 
For ScanRefer, we follow \cite{scanrefer} to use detector-generated object proposals and report the results as Acc@IoU$> k$. 
As shown in Table \ref{vg1}, \model obtains an overall accuracy of 67.5\% and 78.6\% on Nr3D and Sr3D, which has gains of +1.1\% and +1.5\% compared to the previous leading methods.
It indicates that \model is trained on our large-scale \dataset with diverse 3D vision and text can understand complex 3D scenes better.
Notably, our approach can be adaptive to different views better(+1.7\% and +5.2\% on ``View-Dep'' compared with 3D-VISTA), benefiting from the well-explored view-aggregated region-word alignment during the pre-training stage.
Our \model obtains an overall accuracy of 52.3\% @0.25 and 46.2\% @0.5, which outperforms all other methods on ScanRefer.
Compared to the pre-training methods~\cite{jin2023context,zhu20233d}, we introduce more fine-grained pre-training tasks benefiting from the scale-up data and rich and accurate annotations of our \dataset.

\paragraph{3D Dense Captioning.} 
In Table \ref{caption}, we evaluate our model on the Scan2cap.
Our method outperforms the second best scores on CIDEr@0.25 (+1.1\%), METEOR@0.25 (+0.5\%), BLEU-4@0.5 (+2.6\%), METEOR@0.5 (+0.9\%) and ROUGE@0.5 (+0.5\%). 
While our method achieves state-of-the-art in most metrics, it is only comparable with previous methods on BLEU-4@0.25, ROUGE@0.25 and  CIDEr@0.5. The cause could be attributed to the fact that the textual descriptions utilized in pretraining are automatically generated with minimal manual annotations. This may result in gaps and biases about fluency and free-form expression.

\paragraph{3D Question Answering.} 
We evaluate our model on the ScanQA dataset~\cite{scanqa}. 
The quantitative comparison (Table \ref{tab:scanqa}) shows our method surpasses existing models on both EM@1 (+0.6\%/+1.1\%), EM@10(+0.4\%/+0.4\%) and captioning metric ROUGE(+0.6\%/+0.5\%).
While the textual input of 3D-QA is a question rather than referring expression , our method still benefits the 3D QA task with the multi-modal features from multi-grained alignment learning.

\subsection{Ablation Studies} 
We conduct the ablation studies on three tasks in  \Cref{tab:pretrain,tab:domain,tab:level}. `VG' stands for overall accuracy of 3D visual grounding on Nr3D, `Cap' for the CIDEr@0.25 result on Scan2Cap and `QA' for the EM@1 results(w/ objects) on ScanQA.

\begin{table}
\centering
\small

    \begin{tabularx}{\linewidth}{YYYYYY}
    \toprule
    ORP            & VRWA            & MRWA          & VG &Cap& QA \\ \hline
    $\times$    &  $\times$ & $\times$& 60.5        & 63.0 & 25.1     \\
    \checkmark    &  $\times$ & $\times$ & 62.9        &  65.8&25.9       \\
     $\times$    &  \checkmark& $\times$& 63.3      & 67.1 &  26.2   \\  
     $\times$    &   $\times$&\checkmark&  63.1    & 66.8 & 26.5  \\  
    \checkmark    &  \checkmark   & $\times$ &  66.7       &  69.8  & 27.1    \\

    \checkmark    &  \checkmark    & \checkmark   & \textbf{67.5}         & \textbf{72.1} & \textbf{27.6}      \\ 
    \bottomrule
    \end{tabularx}
        \caption{Ablation on Fine-grained Pre-training Tasks. }\label{tab:pretrain}
\end{table}

\begin{table}
\centering
  \small
    \begin{tabularx}{\linewidth}{YYYYYY}
    \toprule
    vision            & lang            &  joint          & VG &Cap& QA \\ \hline
    $\times$    &  $\times$ & $\times$ & 53.1        & 58.7& 23.7      \\
    \checkmark    &  $\times$ & $\times$ & 62.3        & 65.4& 24.5       \\

    \checkmark    &  \checkmark   & $\times$ & 65.1        & 68.1&25.5      \\

    \checkmark    &  \checkmark    & \checkmark   & \textbf{67.5}         & \textbf{72.1} & \textbf{27.6}      \\ 
    \bottomrule
    \end{tabularx}
      \caption{Ablation of  Synthetic-to-Realistic Adaptation. }\label{tab:domain}
\end{table}

{
\paragraph{Effectiveness of the fine-grained pre-training tasks.}
We evaluate the effectiveness of fine-grained alignment tasks in enhancing 3D visual grounding, dense captioning, and question-answering tasks (Table \ref{tab:pretrain}):
(1) \textit{ORP} (Row 2): Compared to the baseline (Row 1), ORP achieves improvements of +2.4\%, +2.8\%, and +0.8\% for 3D visual grounding, dense captioning, and question-answering tasks, respectively. This suggests ORP's effectiveness in understanding object-environment relations in 3D scenes.
(2) \textit{VRWA} (Row 3): VRWA leads to gains of +2.8\%, +4.1\%, and +1.1\%. Its focus on viewpoint-based region-word alignment seems to enhance adaptability from various perspectives.
(3) \textit{MRWA} (Row 4): Incorporating MRWA results in +2.6\%, +3.8\%, and +1.4\% improvements. MRWA's multi-granularity alignment (object/room/scene) appears to enhance model perception in 3D environments.
(4) \textit{VRWA and MRWA v.s. ORP}: Both VRWA and MRWA show more significant improvements in visual grounding and question-answering than ORP, likely due to the enhanced perception across multiple perspectives and granularities from VRWA and MRWA, that is required by VG and QA tasks.
(5) \textit{Integrating All Tasks}: The combination of ORP, VRWA, and MRWA yields the highest gains: +7.0\%, +9.1\%, and +2.5\%, demonstrating their capabilities.
}
{
\paragraph{Effectiveness of synthetic-to-realistic domain adaptation.}
We evaluated the effectiveness of utilising three specific domain adaptation discriminators that address the domain gap between synthetic and real-world down-stream data (Table \ref{tab:domain}).
(1) \textit{3D Vision Domain Adaptation} (Row 2): By applying adaptation to the 3D vision domain, we observe performance increases of 9.2\%, 6.7\%, and 0.8\% for 3D visual grounding, dense captioning, and question-answering tasks, respectively. This highlights the significance of 3D domain adaptation.
(2) \textit{Language Domain Adaptation} (Row 3): Extending adaptation to the language domain further enhances results, with gains of 2.8\%, 2.7\%, and 1.0\%. This step demonstrates the value of refining the language domain.
(3) \textit{Joint Vision-Language Domain Adaptation} (Row 4): Finally, incorporating vision-language joint domain adaptation leads to additional improvements of 2.4\%, 4.0\%, and 2.1\%. This confirms the effectiveness of our comprehensive approach in reducing domain discrepancies and improving overall performance.
}

\begin{table}
\centering
    \small

    \begin{tabularx}{\linewidth}{YYYYYY}
    \toprule
    Object& Room & House        & VG &Cap& QA \\ \hline
    \checkmark    &  $\times$ & $\times$& 66.7       & 69.8& 27.1     \\
    \checkmark    &  \checkmark& $\times$& 67.1     & 71.0& 27.5    \\
     \checkmark& $\times$& \checkmark    &  66.9      & 70.6& 27.3    \\
    \checkmark    &  \checkmark& \checkmark &\textbf{67.5 }        & \textbf{72.1} & \textbf{27.6}     \\
    \bottomrule
    \end{tabularx}
    \caption{Ablation of Multi-level Region-Word Alignment .}\label{tab:level}
\end{table}

{
\paragraph{Ablation of MRWA levels.}
We analyze the impact of different MRWA levels (object, room and house) across three tasks (Table \ref{tab:level}), by comparing with baseline object-only alignment.
(1) \textit{Room-Level Alignment} (Row 2): Adding room-level alignment to the baseline object-only alignment (Row 1) yields improvements of 0.4\%, 1.2\%, and 0.4\%. This suggests that room-level associations between words and objects enhance the model's understanding of object relationships within rooms.
(2) \textit{House-Level Alignment} (Row 3): Extending alignment to the house level (whole scene) leads to gains of 0.2\%, 0.8\%, and 0.2\%. This level of alignment assists the model in comprehending the entire scene by providing associations across the house.
(3) \textit{Combining All Levels} (Row 4): Combining alignments at both room and house levels results in the most substantial improvements of 0.8\%, 2.3\%, and 0.5\%. It indicates the effectiveness of multi-level alignment in improving the model's 3D scene perception.
}

\section{Conclusion}
\label{sec:conclusion}
To tackle the scarcity of data and concerns regarding data collection cost for the 3D vision-language pretraining, we employ synthetic data generation using 3D simulators and off-the-shelf models and create a substantial and diverse synthetic dataset \dataset.
With the rich annotations in \dataset, we pre-train a model \model aligning 3D and language with diverse auxiliary tasks,  including predicting object relationships, multi-level and view-aggregated region-word alignment. A domain adaptation strategy in  fine-tuning is adapted to downstream tasks. 
Extensive comparisons on  various benchmarks  show the effectiveness of our method. 

\section*{Acknowledgements}
This work was supported by the
grants from the National Natural Science Foundation of
China 62372014.

\bibliographystyle{named}
\bibliography{ijcai24}

\end{document}